\documentclass{article}

\usepackage{microtype}
\usepackage{graphicx}
\usepackage{subfigure}
\usepackage{booktabs} 

\usepackage{hyperref}

\usepackage[accepted]{icml2024}

\usepackage{amsmath}
\usepackage{amssymb}
\usepackage{mathtools}
\usepackage{amsthm}

\usepackage[capitalize,noabbrev]{cleveref}

\usepackage{dsfont}
\usepackage{amsmath}
\usepackage{amsthm}
\usepackage{units}
\usepackage{mathtools}
\usepackage{color}

\usepackage{verbatim}
\usepackage{bm}
\usepackage{multirow}

\theoremstyle{plain}

\theoremstyle{definition}

\theoremstyle{remark}

\usepackage[textsize=tiny]{todonotes}

\icmltitlerunning{Submission and Formatting Instructions for ICML 2024}

\begin{document}

\twocolumn[
\icmltitle{Accelerating PDE Data Generation via \\
Differential Operator Action in Solution Space}

\icmlsetsymbol{equal}{*}


\begin{icmlauthorlist}
\icmlauthor{Huanshuo~Dong}{ustc,mail}
\icmlauthor{Hong~Wang}{ustc}
\icmlauthor{Haoyang~Liu}{ustc}
\icmlauthor{Jian~Luo}{ustc}
\icmlauthor{Jie Wang}{ustc}\textsuperscript{\textdagger}
\end{icmlauthorlist}


\icmlaffiliation{ustc}{University of Science and Technology of China}
\icmlaffiliation{mail}{\textless bingo000@mail.ustc.edu.cn\textgreater }

\icmlcorrespondingauthor{Jie Wang}{jiewangx@ustc.edu.cn}


\icmlkeywords{Machine Learning, ICML}

\vskip 0.3in
]



\printAffiliationsAndNotice{\icmlEqualContribution} 

\begin{abstract}
Recent advancements in data-driven approaches, such as Neural Operator (NO), have demonstrated their effectiveness in reducing the solving time of Partial Differential Equations (PDEs).
However, one major challenge faced by these approaches is the requirement for a large amount of high-precision training data, which needs significant computational costs during the generation process.
To address this challenge, we propose a novel PDE dataset generation algorithm, namely \textbf{\underline{Diff}}erential \textbf{\underline{O}}perator \textbf{\underline{A}}ction in \textbf{\underline{S}}olution space (\textbf{DiffOAS}), which speeds up the data generation process and enhances the precision of the generated data simultaneously.
Specifically, DiffOAS obtains a few basic PDE solutions and then combines them to get solutions. It applies differential operators on these solutions,  a process we call 'operator action', to efficiently generate precise PDE data points.
Theoretical analysis shows that the time complexity of DiffOAS method is one order lower than the existing generation method.
Experimental results show that DiffOAS accelerates the generation of large-scale datasets with 10,000 instances by 300 times.
Even with just 5\% of the generation time, NO trained on the data generated by DiffOAS exhibits comparable performance to that using the existing generation method, which highlights the efficiency of DiffOAS.
\end{abstract}

\section{Introduction}
The Partial Differential Equation (PDE) is a fundamental mathematical model derived from various scientific areas including physics, chemistry, biology, engineering and so on~\citep{zachmanoglou1986introduction}.
\begin{figure}[H]
\vskip 0in
\begin{center}
\begin{subfigure}
    \centering
    \raisebox{1cm}{\includegraphics[width=\linewidth]{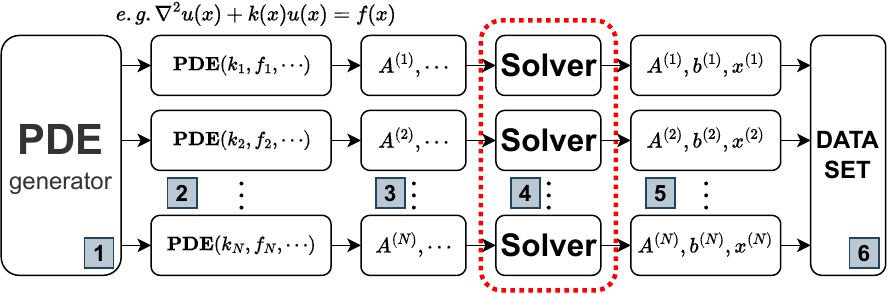}}
    \label{flow chart}
\end{subfigure}
\vskip -0.4in
\begin{subfigure}
    \centering
    \includegraphics[width=\linewidth]{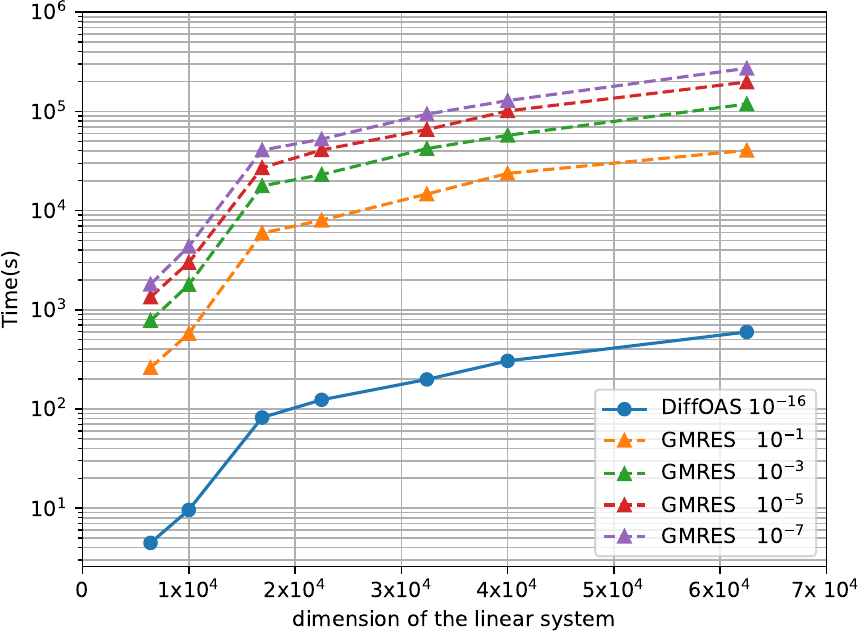}
    \label{figure table}
\end{subfigure}
\vskip -0.1in
\caption{
\textbf{Above.} Generation process of the PDE dataset. 1. Produce collection of random parameters derived from PDE. 2. Generate the relevant PDE using these parameters 3. Convert the PDE into linear equation systems using discretization methods. 4. Solve linear equations based on input parameters 5. Acquire solutions for the linear systems and translate them into solutions for the PDEs. 6. Compile the data into a dataset. \textbf{Below.} The generation cost of DiffOAS and GMRES varies with the dimension of the linear system. GMRES curves in the graph represent different truncation errors, where GMRES $10^{-5}$ indicates the algorithm's time cost with a truncation error of $10^{-5}$. In contrast, our DiffOAS maintains machine precision of $10^{-16}$. It can be observed that DiffOAS significantly speeds up the data generation process, achieving a speedup of up to $70,000$ times.
}
\end{center}
\vskip -0.25in
\end{figure}

Traditionally, solving PDEs often relies on extensive domain expertise and computationally intensive methods.
To reduce the solving time for PDEs, recent research has explored data-driven approaches to predict PDE solutions.
One such approach is the use of Neural Operators  (NOs)~\citep{lu2019deeponet}, which has achieved promising results in accelerating the solving process~\citep{zhang2023artificial}. 

However, the long running time and computation costs of generating the training datasets pose a great challenge to the training of NOs.
Firstly, real-world applications often involve various types of PDEs, and the training of NO for a certain type of PDEs requires a large number of training instances.
For example, when training a Fourier Neural Operator (FNO)~\citep{li2020fourier} for Darcy flow problems, it is common to require thousands of PDEs and their corresponding solutions under diverse initial conditions.
Secondly, acquiring the corresponding solution functions of PDEs as labels for training the NO network poses a further obstacle.
Existing algorithms typically rely on traditional PDE solvers, such as the finite difference method (FDM)~\citep{strikwerda2004finite} and finite element method (FEM)~\citep{leveque2002finite}.
These algorithms, demonstrated in Figure \ref{flow chart}, involve solving large-scale linear equation systems in the \(4\)-th step \textit{Solver}, with a high computational complexity of $O(n^3)$, where $n$ represents the dimension of the linear system.
Consequently, utilizing such algorithms can be time-consuming, with the computational cost of the \textit{Solver} module often accounting for over 95\% of the entire data generation process~\citep{hughes2012finite}.
Thirdly, solving large-scale linear systems often involves the use of iterative methods. However, due to the presence of termination conditions, these methods inevitably introduce errors, which can potentially result in a degradation of the performance of the NO network. Furthermore, as shown in Figure~\ref{figure table}, increasing the accuracy of solving linear systems leads to increased time costs.
Therefore, these challenges of data generation have significantly hindered the real-world applications of NOs~\citep{zhang2023artificial}.




To address the aforementioned challenges, we propose a novel and efficient PDE data generation method, named \textbf{\underline{Diff}}erential \textbf{\underline{O}}perator \textbf{\underline{A}}ction in \textbf{\underline{S}}olution space (\textbf{DiffOAS}).
DiffOAS has two key advantages: it accelerates the data generation process and enhances the precision of the generated data simultaneously.
DiffOAS replaces the process of solving linear systems in the \textit{Solver} module in Figure \ref{flow chart} with operator actions. Initially, we generate a set of PDE solution functions that comply with the actual physical contexts, which serve as basis functions for the solution space. These basis functions are then appropriately combined to satisfy the PDE conditions and generate solution functions. The discretized operator and solution functions are used as inputs in the \textit{Solver}, where the corresponding differential operators are applied to calculate the remaining data that satisfies PDE constraints.
DiffOAS utilizes operator actions to avoid the process of solving linear equation systems, reducing the computational complexity by one order. As shown in Figure~\ref{figure table}, the DiffOAS method can significantly accelerate PDE data generation, achieving a speedup of up to 300 times.





The distinct contributions of our work can be summarized as follows.
\begin{itemize}
    \item We introduce a novel PDE dataset generation algorithm that utilizes differential operator actions. This algorithm generates a set of PDE solution functions that align with the physical background. By applying differential operators to combinations of these solution functions, it can generate large-scale PDE data.
    \item We have demonstrated in our theoretical analysis that our proposed algorithm is able to achieve mechanical precision at a low cost compared to existing generation methods, ensuring the accuracy of generated data
    \item We demonstrate that our proposed algorithm significantly reduces the computational complexity and data generation time, which addresses the long-standing challenge of the data-driven approaches for solving PDEs.
    Even with just 5\% of the generation time, NO trained on the data generated by DiffOAS exhibits comparable performance to that using existing generation methods.
\end{itemize}

\section{Related Work}

\subsection{Data-Efficient and learning-based PDE Algorithms}

Data-efficient and learning-based algorithms have significantly impacted the realm of PDEs. Major advancements include the development of NOs such as the Fourier Neural Operator (FNO)~\citep{li2020fourier} and the Deep Operator Network (DeepONet)~\citep{lu2019deeponet}, both of which have considerably advanced the solving of PDEs. These models harness deep learning to unravel the complexities inherent in PDE systems, providing efficient alternatives to conventional methods.
Additionally, there are also studies exploring the use of neural networks to accelerate the solution of linear equation systems, thereby speeding up the process of solving PDEs. The evolution of data-driven solvers, exemplified by the data-optimized iterative schemes in studies such as \citet{hsieh2019learning, yang2016data, li2023learning, kaneda2023deep}, highlights a trend towards merging machine learning with traditional computational techniques to enhance algorithmic efficiency.


\subsection{Data Generation for PDE Algorithms}



Training data-driven PDE algorithms requires large offline paired parametrized PDE datasets. Typically, the generation of PDE datasets is obtained by solving PDE problems employing traditional computational mathematics algorithms.
In the field of computational mathematics, the numerical solution of complex PDEs generally involves converting the intricate equations into solvable linear systems using various discretization methods~\citep{morton2005numerical}, such as Finite Difference Method (FDM)~\citep{strikwerda2004finite}, Finite Element Method (FEM)~\citep{hughes2012finite, johnson2012numerical}, and Finite Volume Method (FVM)~\citep{leveque2002finite}. These approaches ultimately result in the formation of large linear equation systems, which are usually solved using iterative methods~\citep{liesen2013krylov} suited to the matrix properties~\citep{golub2013matrix}, such as the Conjugate Gradient (CG) algorithm for SPD matrices~\citep{hestenes1952methods}, the Minimum Residual (MINRES) Method for symmetric matrices~\citep{paige1975solution}, and the Generalized Minimum Residual (GMRES) Method for nonsymmetric matrices~\citep{saad1986gmres}.

Although these traditional methods are effective, they are not exclusively designed for dataset generation, and using them independently for this purpose results in substantial computational costs. This has emerged as a significant barrier to the further advancement of data-driven approaches~\citep{zhang2023artificial, hao2022physics}. In tackling this challenge, research has led to the development of architectures that preserve symmetries and conservation laws~\citep{brandstetter2022clifford, liu2023ino, wang2024accelerating}, enhancing model generalization and data efficiency. However, these advancements mainly focus on the optimization of the PDE solving algorithms themselves, without significantly altering data generation methods.

\section{Preliminaries}

\subsection{Discretization for PDEs}\label{Discretization}

Our main focus is the generation of PDE datasets, which are obtained by solving relevant PDE problems. Due to the complexity, continuity, and detailed boundary conditions of these PDE problems, discretized numerical methods such as FDM, FEM, and FVM are typically used to solve them~\citep{strikwerda2004finite, hughes2012finite, johnson2012numerical, leveque2002finite, cheng2023solving}.



Numerical methods discretize a partial differential equation problem by mapping it from an infinite-dimensional function space to a finite-dimensional space, resulting in a system of linear equations. 
To illustrate, we consider solving a two-dimensional inhomogeneous Helmholtz equation using the FDM, which transforms it into a linear equation system:
\begin{equation}
    \nabla^2 u(x, y) + ku(x, y) = f(x, y),
	\label{eq:i_Helmholtz}
\end{equation}
using a \( 2 \times 2 \) internal grid (i.e., \( N_x = N_y = 2 \) and \( \Delta x = \Delta y \)), the unknowns \( u_{i,j} \) can be arranged in row-major order as follows: \( u_{1,1}, u_{1,2}, u_{2,1}, u_{2,2} \). For central differencing on a \( 2 \times 2 \) grid, the vector \( \bm{b} \) will contain the values of \( f_{i,j}=f(x_i, y_j) \) and the linear equation system \( \bm{A} \bm{x} = \bm{b} \) can be expressed as:
\[
\begin{bmatrix}
-4+k & 1 & 1 & 0 \\
1 & -4+k & 0 & 1 \\
1 & 0 & -4+k & 1 \\
0 & 1 & 1 & -4+k
\end{bmatrix}
\begin{bmatrix}
u_{1,1} \\
u_{1,2} \\
u_{2,1} \\
u_{2,2}
\end{bmatrix}
=
\begin{bmatrix}
f_{1,1} \\
f_{1,2} \\
f_{2,1} \\
f_{2,2}
\end{bmatrix}
.
\]
By employing various methods to generate function \( f \) and constant \( k \), such as utilizing Gaussian Random Fields (GRF) or Uniform Random Distribution, we can derive inhomogeneous Helmholtz equations characterized by distinct parameters.

This represents a relatively simple PDE problem. However, for realistic physical simulations and more complex boundary conditions, numerically solving PDEs demands more specialized discretization methods and denser grid divisions. This results in a substantial increase in the size of the matrices within the linear systems, with matrix dimensions potentially growing from \(10^3\) to \(10^6\) or even larger.
This results in significant computational costs during the dataset generation process. 

\subsection{Details of the Dataset}\label{Dataset Detail}
To introduce the content of the dataset, we will use a 2D Darcy Flow problem as an example~\citep{li2020fourier}:
$$
\begin{aligned}
\nabla \cdot (a(x,y)\nabla u(x, y)) & = f(x, y)& &(x,y)\in D\\
u(x, y)& =0 & &(x,y)\in \partial D,
\label{eq:dracy flow}
\end{aligned}
$$

where $D = [0,1]^2$. The functions \( a(x,y) \), \( f(x,y) \), and \( u(x,y) \) represent parameter functions, solution functions, and right-hand side term functions. They are discretized on a $N \times N$ uniform grid $\Omega=\{(i/N,j/N),i, j = 0,1,\cdots, N\}$. So, the dimension of the matrix $\bm{A}$ obtained from the discrete PDE is given by $n=N \times N$. According to the given grid $\Omega$, generate a dataset with
features $F_k = (a_k(\Omega),f_k(\Omega))$ and target $T_k=(u_k(\Omega))$, where $k = 1,2,...,N_{samples}$. In existing data generation methods, the solution function \( u(x,y) \) is obtained by solving the equation using \( a(x,y) \) and \( f(x,y) \), which can be constants, generated through GRFs, or other generation methods. 



\section{Method}\label{Method}

\begin{figure}[ht]
\vskip 0.1in
\begin{center}
\centerline{\includegraphics[width=\columnwidth]{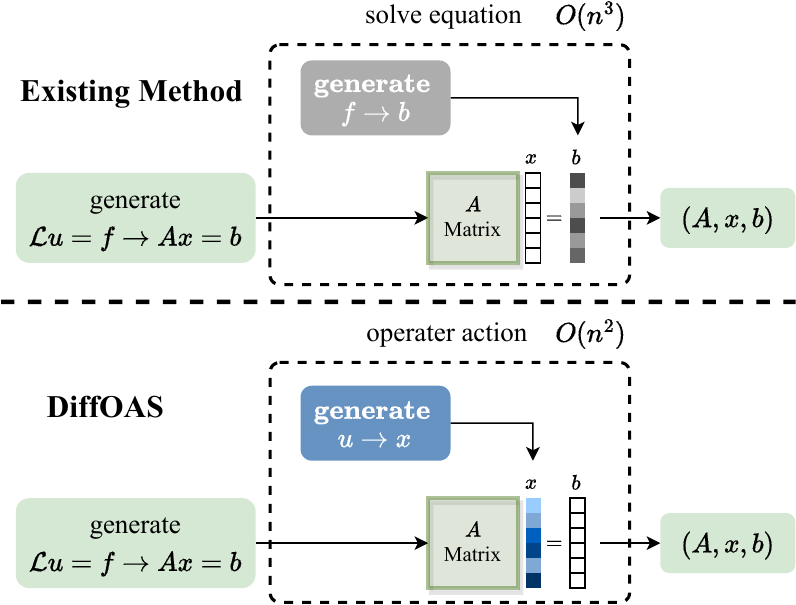}}
\caption{
Overview of the model architecture. 
The process of DiffOAS and existing method: Firstly, Transforming a PDE into a linear equation system.
Secondly, (\textbf{Above}) \textbf{Existing Method} generates $f$, input its discretized form $\bm{b}$ into the linear equation system, and solve $\bm{x}$ accordingly. (\textbf{Below}) \textbf{DiffOAS} generates $u$ using basis functions, inputs its discretized form $\bm{x}$ into the linear equation system, and calculates $\bm{b}$ accordingly.
Finally organizing the inputs and outputs of the linear equation system to create a complete PDE dataset.
}
\label{Method Figure}
\end{center}
\vskip -0.5in
\end{figure}


In this paper, we explore the most prevalent form of PDEs, namely the steady-state PDEs constrained by boundary conditions~\citep{evans2022partial}. These can generally be expressed in the following form:
\begin{equation}
\begin{aligned}
\mathcal{L}(a(\bm{x})) u(\bm{x}) & = f(\bm{x})& & \bm{x}\in D \\
\mathcal{B}(b(\bm{x})) u(\bm{x}) & = g(\bm{x})& & \bm{x}\in \partial D.
    \label{eq:PDE}    
\end{aligned}
\end{equation}
Here, \( D \subseteq \mathbb{C}^n \) is the PDE's domain, with \( \partial D \) as its boundary, and $\bm{x} \in \mathbb{C}^n$. The operator \( \mathcal{L} \), governed by the parameter function \( a(\bm{x}) \), is a partial differential operator. The solution function is \( u(\bm{x}) \), and \( f(\bm{x}) \) represents the source or forcing function on the right-hand side. The boundary operator \( \mathcal{B} \), controlled by \( b(\bm{x}) \), dictates the equation's boundary conditions on \( \partial D \), and \( g(\bm{x}) \) is the corresponding right-hand side function on \( \partial D \).




In existing algorithms, the dataset generation process began with the random generation of other parameters for PDEs, followed by employing traditional PDE solvers to determine the corresponding solution functions \( u(\bm{x}) \). This approach necessitated the \textit{Solver} module in Figure~\ref{flow chart} to solve large linear systems of equations involving substantial matrices, as mentioned in the Preliminaries~\ref{Discretization}. As mentioned in Figure~\ref{figure table}, existing method requires a substantial amount of time and often becomes a bottleneck in the dataset generation process.

As shown in Figure~\ref{Method Figure}, unlike traditional methods, DiffOAS method initiates by generating solution functions \( u(\bm{x}) \). Subsequently, these functions are subjected to operator actions to derive other PDE parameters, such as \( f(\bm{x}) \), thereby facilitating dataset creation. Both methods generate data that adhere to their respective PDE constraints. However, our algorithm circumvents the significant computational expense and termination error associated with solving large matrix linear systems. This strategy substantially reduces computational overhead and enhances precision.

DiffOAS method comprises two primary steps:

1. \textbf{Solution Functions Generation}: This phase involves the low-cost generation of solution functions that comply with PDE conditions.

2. \textbf{Operator Action}: In this stage, operators are applied to the generated solution functions to obtain data that satisfies PDE constraints.


\subsection{Solution Functions Generation}

DiffOAS method specifically generates a series of basis functions either through a designated distribution that satisfies actual physical contexts (typically choosing between 10 to 50, taking $N_{basis}$ as an example). This paper utilizes \textbf{existing method} to generate these basis functions. These basis functions form the foundational elements in the solution space for that particular physical distribution. We randomly weight and normalize these basis functions using Gaussian distribution. Additionally, we introduce a noise element \(\epsilon\) that maintains the boundary conditions unaltered (further details are provided in the Appendix~\ref{data set}), resulting in a novel solution function \(u_{new}(\bm{x})\).
\begin{align*}
u_{\text{new}} = \sum_{i=1}^{N_{basis}}\alpha_{i} u_i+\epsilon& &
\alpha_{i}=\frac{\mu_i}{\sum_{j=1}^{N_{basis}}\mu_j}\\
\mu_i \sim N(0,1)& & i=1,2,...,N_{basis}.
\end{align*}
This method of weighting ensures that the newly formulated solution function \(u_{new}(\bm{x})\) complies with the boundary conditions of the PDE, while also preserving its physical relevance.
The incorporation of noise serves to enhance the complexity and applicability of the generated dataset. Through this methodology, we are capable of producing a myriad of physically meaningful solution functions at minimal expense, utilizing a set of basis functions that align with a physical distribution.

\subsection{Operator Action}

The operator \(\mathcal{L}\) is a functional mapping within a Sobolev space associated with PDEs. This mapping transforms functions to functions as detailed below:
\begin{equation}
\begin{aligned}
    &\mathcal{L}: \mathcal{U} \rightarrow \mathcal{F}\\
    &\quad : u \mapsto f,
    \label{eq:L}    
\end{aligned}
\end{equation}
where \(\mathcal{U}\) and \(\mathcal{F}\) represent the Sobolev spaces for the solution functions and the right-hand side functions of the PDE, respectively.
Here, \( u \) is a solution function in space \( \mathcal{U} \), and \( f \) is the corresponding right-hand side function in space \( \mathcal{F} \) for the PDE problem~\eqref{eq:PDE}~\citep{evans2022partial}.

Specifically, as mentioned in the Preliminaries~\ref{Discretization} regarding the Helmholtz equation~\eqref{eq:i_Helmholtz}, the Sobolev function spaces are defined as \(\mathcal{U}=\mathcal{F}=\mathcal{W}^{1,2}([0,1]^2) = \mathcal{H}^1([0,1]^2)\), where the norm for \(\mathcal{H}^1([0,1]^2)\) is given by
\begin{equation}
    \|f\|_{\mathcal{H}^1} = \left( \int_{[0,1]^2} |f|^2 + |\nabla f|^2 \right)^{\frac{1}{2}}.
    \label{eq:H1}   
\end{equation}
The operator corresponding to the Equation~\eqref{eq:i_Helmholtz} is \(\mathcal{L} = \nabla^2 + k\), and \(\mathcal{L}u=(\nabla^2 + k)u(\bm{x})\) represents the action of the operator \(\mathcal{L}\) on the solution function \(u(\bm{x})\).

In numerical computations, when applying an operator to a function, we utilize discretization methods for PDEs as discussed in the Preliminaries~\ref{Discretization}. This process essentially projects the PDE from an infinite-dimensional Sobolev function space to a finite-dimensional linear space. It represents the differential operator \( \mathcal{L} \) as a linear transformation \( \bm{A} \), and functions are expressed in vector form: solutions \( u \) as vector \( \bm{x} \), and the right-hand side functions \( f \) as vector \( \bm{b} \). Thus, the PDE problem is transformed into a linear equation system:
\begin{equation}
\begin{aligned}
    \mathcal{L}u=f \rightarrow \bm{A}\bm{x}=\bm{b}.
    \label{eq:linear equation}
\end{aligned}     
\end{equation}
In this context, the existing method used for dataset generation translates into solving large linear equation systems by finding \( \bm{x} \) using \( \bm{A} \) and \( \bm{b} \). However, in our method, the operator action is converted into multiplying matrix \( \bm{A} \) by vector \( \bm{x} \) to obtain \( \bm{b} \).

For the same problem, the computation required for a single matrix-vector multiplication is significantly less than that for solving the corresponding system of linear equations, and it does not introduce additional errors from the solving process. Therefore, our algorithm offers greater speed and higher precision.

\section{Theoretical Analysis}

\subsection{Computational Complexity Analysis}
\label{Complexity 1}
\subsubsection{Existing Method}


Solving the corresponding linear systems constitutes the primary computational expense in numerical PDE solving~\citep{hughes2012finite}. Consequently, the computational cost of generating data points using the existing algorithm can be estimated by the cost of solving these linear systems. The GMRES, an iterative algorithm based on the Krylov subspace~\citep{liesen2013krylov, qin2023solving}, is commonly used for large, sparse, and nonsymmetric linear systems. Iteratively constructing the Krylov space, this method projects the linear equation system of a large matrix onto a smaller subspace, effectively reducing computational complexity. The specific pseudocode is provided in Appendix~\ref{GMRES}.

The most computationally intensive components are the matrix-vector multiplication and the orthogonalization process. Assuming that the matrix dimension is \(n\) and the current iteration number is denoted by \(j = 1, 2, \cdots, m\) with a total of \(m\) iterations,  where \(m\) represents the final dimension of the Krylov subspace.

The complexity of each iteration is primarily determined by \(O(n^2)\) for the matrix-vector product and \(O(j \times n)\) for the orthogonalization process. In the worst case scenario where \(m\) approaches \(n\), the total complexity over \(m\) iterations can be approximated as \(O(m \times n^2)\) for the matrix-vector products and \(O(m^2 \times n)\) for the orthogonalization. The overall computational cost can be approximated by \(O(m \times n^2 + m^2 \times n)\).
Furthermore, in practical scenarios, the computational complexity is influenced by the matrix's sparsity and the algorithm's implementation specifics.

Assuming the dataset contains \(N\) data points, the computational cost of generating the dataset using the existing method can be approximated as \(O(m \times n^2 \times N + m^2 \times n \times N)\).



\subsubsection{DiffOAS Method}




According to the introduction in Method~\ref{Method}, the DiffOAS method is divided into two steps: solution functions generation and operator action. In numerical computations, as explained in Preliminaries~\ref{Discretization}, we use discretization methods to represent PDE operators as matrices and implement operator actions by multiplying these discretized matrices with vectors corresponding to the solution functions.


In the process of generating the solution function, we utilize existing method to generate basis functions and construct the solution function space based on these functions. Since the computational cost of generating the solution function space from the basis functions is negligible, our focus will be solely on discussing the time complexity of generating the basis functions. Let \(l\) denote the number of basis functions, which is typically much smaller than the dataset size 
\(N\) (with \(l\) typically chosen to be between 10 and 50). The time complexity of this step can be expressed as \(O(m \times n^2 \times l + m^2 \times n \times l)\).


The operator action step is a single matrix-vector multiplication operation. The time complexity for dense matrices is \(O(n^2)\). In particular, the time complexity can be influenced by the sparsity of the matrix. If the dataset contains \(N\) data points, the computational cost of generating the dataset using our method is approximately \(O(n^2 \times N)\).


Therefore, the overall computational complexity of our algorithm can be expressed as \(O(n^2 \times N + m \times n^2 \times l + m^2 \times n \times l)\). By comparing the computational complexities of the existing method and our method, since \(l\) is a constant value and considerably smaller than \(N\). \(m\) is mathematically of the same order as \(n\) (typically ranging between \(n/20\) to \(n/5\) in experiments). Consequently, our method theoretically provides a speedup of approximately \(O(m) \approx O(n)\), which corresponds to an increase in speed by one order, where \(m\) and \(n\) are of the same order.

\subsection{Error Analysis}\label{error analysis}

In constructing datasets for physical problems, errors primarily arise from three sources: 1. Physical Modeling Error, which refers to the discrepancy between the PDE and the actual physical processes; 2. Discretization Error, which occurs due to the chosen numerical PDE methods and grid densities; 3. Data Generation Error, which is associated with the errors in generating solution functions that satisfy the grid constraints. In our analysis, we focus on the third type of error, assuming the accuracy of the PDE and fixed discretization methods.


Inaccuracies in solving linear systems are the main source of error in data generation using existing algorithms. 
For Krylov subspace algorithms, such as GMRES, to solve large sparse nonsymmetric linear systems, the magnitude of \( h_{j+1,j} \) from the matrix \( \bm{H}_j \) (generated during the Krylov subspace iteration) serves as a termination criterion for the iterations.

The GMRES algorithm is used to solve linear systems of the form \(\bm{A}\bm{x}=\bm{b} \) by minimizing the residual norm \(\|\bm{A}\bm{x}-\bm{b}\|\). This is achieved through the construction of a Krylov subspace and the utilization of the Arnoldi process to generate an orthogonal basis for the subspace.

During the \( j \)-th iteration of the Arnoldi process, a Hessenberg matrix \( \bm{H}_j \) and an orthogonal matrix \( \bm{V}_{j+1} \) are generated, satisfying the equation \( \bm{A}\bm{V}_j = \bm{V}_{j+1}\underline{\bm{H}}_j \). In this context, \( \bm{V}_j \) contains the basis vectors of the Krylov subspace, and the augmented Hessenberg matrix \( \underline{\bm{H}}_j \) includes an additional row \( {h}_{j+1,j} \).

The residual of the \( j \)-th iteration's approximate solution \( \bm{x}_j \), denoted as \( \bm{r}_j = \bm{b} - \bm{A}\bm{x}_j \), can be expressed as \( \bm{r}_j = \bm{V}_{j+1}(\beta \bm{e}_1 - \underline{\bm{H}}_j \tilde{\bm{y}}) \). Here, \( \beta = \|\bm{r}_0\| \) represents the norm of the initial residual, and \(\bm{e}_1\) denotes the first unit vector in the standard basis of the corresponding space. The vector \( \tilde{\bm{y}} \) an augmented version of \( \bm{y} \), is obtained by appending an additional zero element to the solution vector \( \bm{y} \), which is determined by the central minimization problem in GMRES.

The norm of the residual \( \|\bm{r}_j\| \) can be expressed as  \( \|\bm{r}_j\| = \|\beta \bm{e}_1 - \underline{\bm{H}}_j \tilde{\bm{y}}\| \). By applying the triangle inequality, an estimation can be derived
\begin{equation}
    \|\bm{r}_j\| \leq \|\beta \bm{e}_1 - \bm{H}_j \bm{y}\| + \|h_{j+1,j}y_j \bm{e}_{j+1}\|.
    \label{GMRES_error}
\end{equation}

Given that \( \|\beta \bm{e}_1 - \bm{H}_j \bm{y}\| \) is minimized through the iterative process, the major component of the residual arises from \( \|h_{j+1,j}y_j \bm{e}_{j+1}\| \), which aids in the approximation
\begin{equation}
    \|\bm{r}_j\| \leq |h_{j+1,j}|\|y_j\|. 
    \label{GMRES_error_ap}
\end{equation}
The magnitude of \( h_{j+1,j} \) in GMRES algorithm serves as an indicator for estimating the upper bound of the error in \( \bm{r}_j \). A smaller value of \( h_{j+1,j} \) typically indicates a reduced residual, suggesting a close approximation of \( \bm{x}_j \) to the actual solution.

The computational cost of solving linear systems is directly influenced by the predefined error. We utilize the upper bound equation for error, Equation~(\ref{GMRES_error_ap}) to establish a requirement for the magnitude of \( {h}_{j+1,j} \), continuing iterations when it is substantial and terminating them if it falls below a threshold. The precision requirement for the dataset varies depending on the inherent accuracy of the data-driven algorithm. For example, in algorithms like FNO, where the final error typically ranges from \(10^{-2}\) to \(10^{-4}\), the relative error in the data should ideally be around \(10^{-7}\) to \(10^{-10}\) or lower to maintain the significance of training process.


In our method, the operator is actioned to the generated solution functions, essentially performing matrix-vector multiplication. The precision of this operation is governed by the machine epsilon of floating-point arithmetic in computers, typically, the error is around \(10^{-16}\) to \(10^{-17}\), or even lower. Achieving this high level of precision is nearly impossible with existing method.

To demonstrate how larger data errors can impact model training, we conducted additional experiments detailed in Appendix~\ref{Error Experiments}, where we trained models using datasets with varying error levels.

\section{Experiment}
In this chapter, we compare our proposed data generation method with existing data generation methods.
\subsection{Experimental Setup}
 Our analysis examines three main performance indicators, which are crucial for evaluating the effectiveness of data generation methods:
\begin{itemize}
    \item Accuracy of the data.
    \item Time cost of generating data.
    \item Errors obtained from training on neural operator models.
\end{itemize}
 In our experiments, we focus on testing two widely recognized and widely used neural operator models, which are the most prominent and common models in data-driven algorithms for PDE:
 \begin{itemize}
     \item FNO (Fourier Neural Operator) \citep{li2020fourier}
     \item DeepONet (Deep Operator Network) \citep{lu2019deeponet}
 \end{itemize}

\begin{table*}[h]
\centering
\caption{
Compare the data generation time and training results on different models between our DiffOAS method and GMRES methods. The first row lists the method used to generate the dataset and the number of training instances. The first column represents the corresponding PDE problem and the corresponding length of the matrix sides. Bolding indicates that our algorithm outperforms existing method. 
}
\label{tab:accuracy}
\vskip 0.15in
\begin{center}
\begin{scriptsize}
\begin{sc}
\renewcommand{\arraystretch}{1.6}
\setlength{\tabcolsep}{8pt} 
\begin{tabular}{lccccccccccc}
\toprule
\multicolumn{1}{c}{\multirow{2}{*}{\textbf{Dataset}}} & \multicolumn{3}{c}{\textbf{DARCY FLOW 10000 }} & \quad & \multicolumn{3}{c}{\textbf{WAVE  22500}} & \quad & \multicolumn{3}{c}{\textbf{ DIFFUSION 62500}} \\ \cline{2-4} \cline{6-8} \cline{10-12}
\multicolumn{1}{c}{} & TIME$(s)$ & FNO & DeepONet & \quad  & TIME$(s)$ & FNO & DeepONet & \quad & TIME$(s)$ & FNO & DeepONet \\[0pt] \midrule
GMRES \ \ 1000       &2.99e2  &4.56e-3           & 6.82e-2             &\quad   &3.27e3  & 4.05e-4  & 6.12e-2            &\quad  &1.99e4  &4.29e-3   & 7.70e-2    \\
DiffOAS 1000     &8.97e0  &2.15e-2           & \textbf{6.82e-2}    &\quad   &1.60e2  & 1.71e-3  & \textbf{4.57e-2}   &\quad  &5.99e2  &1.76e-2   &\textbf{6.40e-2}   \\
DiffOAS 5000     &9.20e0  &6.86e-3           & \textbf{5.58e-2}    &\quad   &1.60e2  &7.25e-4   & \textbf{4.74e-2}   &\quad  &6.01e2  &5.66e-3   & \textbf{5.80e-2}     \\ 
DiffOAS 10000    &9.49e0  &\textbf{4.26e-3}  & \textbf{5.62e-2}    &\quad   &1.60e2  &6.52e-4   & \textbf{4.72e-2}    &\quad  &6.03e2     & 4.79e-3  & \textbf{6.04e-2}    \\ \bottomrule
\end{tabular}
\end{sc}
\end{scriptsize}
\end{center}
\vskip -0.1in
\end{table*}

 We tested three different types of PDE problems that have important applications in science and engineering ( detailed descriptions are listed in the Appendix~\ref{specific experimental details}:
 \begin{itemize}
     \item Darcy Flow Problem~\citep{li2020fourier}
     \item Scalar Wave Equation in Electromagnetism~\citep{zhang2022hybrid}
     \item Solute Diffusion in Porous Media~\citep{mauri1991dispersion}
 \end{itemize}


\textbf{Baselines}. The main time expense of existing data generation methods is solving a system of linear equations composed of large sparse nonsymmetric matrices~\citep{hughes2012finite}. We use an existing data generation method based on the \textbf{GMRES} algorithm as the solution and baseline for our study, utilizing scipy 1.11.4~\cite{2020SciPy-NMeth}. For detailed GMRES algorithmic information, please refer to Appendix~\ref{GMRES}.

For constructing the FNO and DeepONet models, we employed 100 instances of test data generated using GMRES methods, The detailed settings are presented in Appendix~\ref{model_set}. The data generation process was performed on Intel(R) Xeon(R) Gold 6154 CPU @ 3.00GHz, while the model training took place on a GeForce RTX 3090 GPU with 24GB of memory.

\subsection{Training Result}

The main experimental results for all datasets and models are shown in Table~\ref{tab:accuracy}. More details and hyperparameters can be found in Appendix~\ref{specific experimental details}. Based on these results, we have the following observations.

Firstly, DiffOAS method consistently demonstrates remarkable acceleration compared to GMRES methods across datasets of all sizes, particularly in the case of Darcy Flow where the acceleration ratio can reach approximately $30$ times. Furthermore, the time required for DiffOAS to generate different numbers of training instances remains relatively constant.
The time required for our method to generate data can be divided into two parts: generating basis functions and applying operators. The time to generate basis functions depends on the number of basis functions and is independent of the number of training instances generated. The time to apply operators is directly proportional to the number of training instances generated. Therefore, the experimental results indicate that the generation of basis functions constitutes the main portion of the data generation time for our method. This implies that the DiffOAS method can generate a large amount of training data at a low cost, demonstrating the efficiency of our approach. 


Secondly, in different PDE problems and with different neural operators, the dataset generated by DiffOAS method exhibits better performance compared to GMRES methods on DeepONet. By increasing the training set size, the error can reach the same order of magnitude on FNO. This indicates that our method, while accelerating data generation by $30$ times, can achieve comparable performance to the data generated by existing method. Regarding the phenomenon of slightly higher errors in some problems compared to existing method, it could be attributed to the fact that the test set is obtained through numerical solutions of linear equations, which inherently have larger errors. In contrast, our method generates datasets with machine precision. Therefore, the larger errors in the predictions of models trained on our datasets on the test set could be attributed to errors in the test set itself and discrepancies between the test set and the exact solution.

\subsection{Data Generation Time And Accuracy}

\begin{table}
\centering
\caption{
Compare the data generation time of our DiffOAS algorithm and GMRES with various accuracies, where the error of the DiffOAS algorithm is the machine precision $1\mathrm{E}-16$. 'TIME1' and 'TIME2' represent the total time and operator action time for data generation, respectively. The remaining parameters in the second row indicate the errors of the GMRES algorithm. 'SIZE' indicates the matrix size of the PDE problem.
}
\label{tab:time and accuracy}
\vskip 0.1in
\begin{center}
\begin{scriptsize}
\begin{sc}
\renewcommand{\arraystretch}{1.3}
\setlength{\tabcolsep}{4pt} 
\begin{tabular}{cccccccc}
\toprule
\multirow{2}{*}{size} &\multicolumn{2}{c}{DiffOAS}       & \quad & \multicolumn{4}{c}{ GMRES }   \\ \cline{2-3} \cline{5-8}
\multicolumn{1}{c}{}  & time1     & time2   & \quad  & 1e-1     & 1e-3     & 1e-5    & 1e-7 \\  \hline
2500      &9.732e-1          &2.687e-1       & \quad     & 4.530e1         & 1.394e2          & 2.349e2        &  3.265e2               \\ 
10000     &9.582e0          &5.931e-1        & \quad    & 5.750e2         & 1.790e3          & 2.996e3        &  4.350e3               \\ 
16900     &8.192e1            &8.066e-1      & \quad      & 5.933e3          & 1.771e4           & 2.704e4         &  4.066e4                \\ 
22500     & 1.236e2             &1.004e0     & \quad       & 7.994e3          & 2.305e4           & 4.086e4         &  5.251e4                \\
32400     & 1.983e2            &1.462e0      & \quad       & 1.466e4         & 4.216e4           & 6.561e4         &  9.384e4               \\
40000     &3.051e2              &1.943e0     & \quad      & 2.371e4           & 5.739e4           & 1.010e5        &  1.285e5               \\
62500     &5.971e2              &3.815e0     & \quad        & 4.052e4           & 1.186e5          & 1.977e5        &  2.722e5               \\\bottomrule
\end{tabular}
\end{sc}
\end{scriptsize}
\end{center}
\vskip -0.1in
\end{table}

The high generation speed and accuracy are the main advantages of our method. We tested the data generation time for generating 10,000 data points at different accuracies in seconds. The results are shown in Table~\ref{tab:time and accuracy}. Further experimental details can be found in Appendix~\ref{time test}.

The experimental results show that compared to the GMRES method, the DiffOAS method achieves a speedup of approximately $300$ times in terms of the total time, while the time for operator actions can be accelerated by up to approximately $70,000$ times for a matrix size of $62,500$. Since the generation of basis functions in our method does not depend on the training data size, as we generate sufficiently large data, the speedup ratio will approach the acceleration ratio of the operator action part, as predicted by the theoretical analysis in Section~\ref{Complexity 1}. In other words, the DiffOAS method reduces the time for generating the dataset by one order, significantly enhancing the efficiency of data generation.



In the process of solving linear systems using the GMRES algorithm, as the accuracy requirement increases, the solution time significantly increases. When the accuracy of the GMRES algorithm is improved from $1\mathrm{E}-5$ to $1\mathrm{E}{-7}$, the time increases by approximately 50\%. In contrast, our DiffOAS method achieves a precision of $1\mathrm{E}{-16}$. This indicates that improving the accuracy of the GMRES algorithm comes with expensive computational costs, while our algorithm guarantees data accuracy at machine precision through operator actions. Therefore, the data generated by the DiffOAS algorithm has much higher precision compared to the data generated by existing method.

\begin{table}
\centering
\caption{
For each precision of the GMRES algorithm, a linear regression is performed on the ratio of the matrix size and the acceleration achieved by the DiffOAS method operator. The slope represents the slope of the fitted line, and Pearson's r is the Pearson correlation coefficient, where a value closer to $1$ indicates a more reliable fit.
}
\label{tab:least square method}
\vskip 0in
\begin{center}
\begin{scriptsize}
\begin{sc}
\renewcommand{\arraystretch}{1.3}
\setlength{\tabcolsep}{20pt} 
\fontsize{9}{10}\selectfont
\begin{tabular}{ccc}
\toprule
Accuracy      & slope   & Pearson r    \\ \hline
1e-1          &0.079    &   0.715      \\ 
1e-3          &0.206    &   0.890      \\ 
1e-5          &0.380    &   0.867      \\ 
1e-7          &0.473    &   0.927      \\ \bottomrule
\end{tabular}
\end{sc}
\end{scriptsize}
\end{center}
\vskip -0.1in
\end{table}

\begin{table}
\centering
\caption{
Experimental results comparing datasets generated using different basis functions.}
\label{tab:Ablation Result}
\vskip 0.15in
\begin{center}
\begin{scriptsize}
\begin{sc}
\renewcommand{\arraystretch}{1.3}
\setlength{\tabcolsep}{17pt} 
\fontsize{9}{10}\selectfont
\begin{tabular}{lcc}
\toprule
Basis      & FNO       & DeepONet
\\ \hline
DiffOAS                 &4.260\%    & 5.629\%           \\ 
GRF                    &31.44\%    & 88.08\%           \\ 
Fourier   &93.56\%    & 96.98\%          \\
Chebyshev     &63.11\%    & 68.23\%           \\\bottomrule
\end{tabular}
\end{sc}
\end{scriptsize}
\end{center}
\vskip -0.15in
\end{table}

We analyzed the relationship between acceleration ratio and matrix size in Table~\ref{tab:least square method}. By performing linear regression in the least squares sense on the largest five matrix sizes, we found that Pearson's r reached $0.927$ on the GMRES algorithm with an accuracy of $1\mathrm{E}-7$. This indicates that the partial acceleration ratio due to the operator action in our algorithm is of the same order as N, implying a one order acceleration in our algorithm, consistent with the time Computational Complexity Analysis~\ref{Complexity 1}.

Our method tackles a common scenario: solving systems of linear equations involving large, sparse, non-symmetric matrices. The GMRES algorithm is the mainstream approach for addressing such problems. However, for certain types of equations, specifically when the system is symmetric positive definite, numerical algorithms like MINRES and Conjugate Gradient should also be considered for baseline comparisons. Consequently, we have included comparative experiments with these baselines in Appendix~\ref{Time comparison Experiments}.

\subsection{Ablation Result}

Finally, we conducted an ablation study to demonstrate the significant impact of the solution function generation method in our DiffOAS method on the quality of the dataset. In Table~\ref{tab:Ablation Result}, we tested datasets generated using GRF, Fourier basis, and truncated Chebyshev functions as the basis functions. For specific details about the design of the basis functions, please refer to Appendix~\ref{Ablation}.

Experimental results show that the training results of the data sets corresponding to the solution functions generated by other methods are quite poor on the FNO and DeepONet models. Even the best training results on FNO and DeepONet have errors of more than 30\% and 60\%, respectively. This shows that these data sets have little training value. Therefore, the solution function introduced in the DiffOAS method that conforms to the real physical distribution is necessary as a basis function and is crucial to ensuring the generation of high-quality datasets that can make accurate predictions.




\section{Limitation and Conclusions}

\textbf{Limitation} \ While our approach has shown promise in speeding up PDE dataset generation, there are areas for further exploration: 1. Our accelerated algorithm is designed for general PDE problems with asymmetric coefficient matrices, but it doesn't specifically target particular types of PDE problems. 2. While generating basis functions, we can enhance dataset quality by choosing specific functions from the generated set using optimization techniques.

\textbf{Conclusions} \ In this article, we introduce the DiffOAS algorithm. To our knowledge, this is the first PDE dataset generation method that does not require solving linear equations. Specifically, the algorithm consists of two parts: solution function generation and operator action. By generating solution functions using basis functions, we ensure that the generated data conforms to physical distributions. The operator action significantly accelerates the data generation process while maintaining mechanical precision in our generated data. The DiffOAS method ensures both speed and accuracy, alleviating a significant obstacle to the development of neural networks.

\newpage

\clearpage
\section*{Impact Statements}
In this paper, we propose a new PDE dataset generation method that significantly improves the speed of data generation. This work has great potential in various practical and important scenarios, such as physics, chemistry, biology, and various scientific problems.

\bibliography{example_paper}
\bibliographystyle{icml2024}

\newpage
\appendix
\onecolumn

\section{Specific pseudocode of GMRES}\label{GMRES}
The following computational procedure is adapted from~\citep{golub2013matrix}

\begin{algorithm}
\renewcommand{\algorithmicrequire}{\textbf{Input:}}
\renewcommand{\algorithmicensure}{\textbf{Output:}}
\caption{Generalized Minimal Residual Method (GMRES)}
\begin{algorithmic}[1]\label{GMRES_pcode}
\REQUIRE $\bm{A} \in \mathbb{R}^{n \times n}$, $\bm{b} \in \mathbb{R}^n$,$\bm{x_0}$ is the initial vector, tolerance $\epsilon > 0$, maximum iterations $m$
\ENSURE Approximate solution $\bm{x}$
\STATE $\bm{r}_0 = \bm{b} - \bm{A}\bm{x}_0$ 
\STATE $\beta = \|\bm{r}_0\|_2$ 
\STATE $\bm{v}_1 = \bm{r}_0 / \beta$
\FOR{$j = 1$ to $m$}
    \STATE $\bm{w}_j = \bm{A}\bm{v}_j$
    \FOR{$i = 1$ to $j$}
        \STATE $h_{i,j} = (\bm{w}_j, \bm{v}_i)$ 
        \STATE $\bm{w}_j = \bm{w}_j - h_{i,j}\bm{v}_i$
    \ENDFOR
    \STATE $h_{j+1,j} = \|\bm{w}_j\|_2$
    \IF{$h_{j+1,j} = 0$}
        \STATE \textbf{break}
    \ENDIF
    \STATE $\bm{v}_{j+1} = \bm{w}_j / h_{j+1,j}$
    \STATE Form or update the Hessenberg matrix $\bm{H}_j$
    \STATE Solve the least squares problem: $\min_{\bm{y} \in \mathbb{R}^j} \| \beta \bm{e}_1 - \bm{H}_j \bm{y} \|_2$
    \STATE Update the solution: $\bm{x}_j = \bm{x}_0 + \bm{V}_j \bm{y}$
    \IF{$\|\bm{r}_j\|_2 < \epsilon$} 
        \STATE \textbf{break}
    \ENDIF
\ENDFOR
\STATE $\bm{x} = \bm{x}_{j}$
\end{algorithmic}
\end{algorithm}

\section{Specific Experimental Details}\label{specific experimental details}

\subsection{Model set}\label{model_set}
\textbf{FNO}: We employ $4$ FNO layers with learning rate $0.001$, batch size $20$, epochs $500$, modes $12$, and width $32$.

\textbf{DeepONet}:
We utilize branch layers: [50, 50, 50, 50, 50, 50, 50, 50, 50, 50, 50, 50] and trunk layers: [50, 50, 50, 50, 50, 50, 50, 50, 50, 50, 50, 50], with the activation function set to tanh. The learning rate is $0.001$, batch size is $20$, and the training process is performed for $500$ epochs.

\subsection{Data}\label{data set}

\subsubsection{Darcy Flow}
In this research, we delve into two-dimensional Darcy flows, which are governed by the equation~\citep{li2020fourier, rahman2022u, kovachki2021neural, lu2022comprehensive}:
\begin{equation}
- \nabla \cdot (K(x, y) \nabla h(x, y)) = f(x, y),
\end{equation}
where \(K\) represents the permeability of the medium, \(h\) denotes the hydraulic pressure, and \(f\) is a source term that varies, being either a constant or a function dependent on spatial variables.


For our experimental setup, the permeability field \( K(x, y) \) and the source term \( f(x, y) \) are generated by the Gaussian Random Field (GRF) methodology, with a time constant $\tau=7$ and a decay exponent $\alpha=2.5$.

In DiffOAS method, In the DiffOAS method, we use $30$ solution functions obtained by solving as basis functions.

\subsubsection{Scalar Wave Equation in Electromagnetism}

In this research, we delve into a two-dimensional Helmholtz equation in electromagnetism, expressed as~\citep{zhang2022hybrid}:
\begin{equation}
\Delta \Phi(x,y) + k^2(x,y) \Phi(x,y) = S(x,y),
\end{equation}
where \(\Phi(x,y)\) is the electromagnetic scalar potential and \(k(x,y)\) the spatially varying wavenumber. \(S(x,y)\) is the source term representing electromagnetic wave origins.


This equation is fundamental in modeling electromagnetic wave propagation and interaction in various media. For our experimental setup, \( k^2(x, y) \) and the source term \(S(x,y)\) are generated by the Gaussian Random Field (GRF) methodology, specifically as $GRF(\tau=3,\alpha=2)/10$.

In DiffOAS method, In the DiffOAS method, we use $50$ solution functions obtained by solving as basis functions.




\subsubsection{Solute Diffusion in Porous Media}

We investigate the process of solute diffusion in a two-dimensional porous medium, described by the following equation~\citep{perkins1963review, mauri1991dispersion}:
\begin{equation}
\nabla \cdot (k(x, y) \nabla C(x, y)) + q(x,y) C(x,y) = f(x,y),
\end{equation}
where \(C(x, y)\) symbolizes the solute concentration, \(k(x, y)\) signifies the diffusion coefficient varying spatially, and \(q(x,y) C(x,y)\) represents the influence of internal or external sources/sinks. The function \(f(x,y)\) acts as an additional source or sink, pinpointing regions of solute addition or removal.

This equation is key for modeling solute movement in heterogeneous porous media. For our experimental setup, the diffusion coefficient \( k(x, y) \) is generated by $10*GRF(\tau=3,\alpha=2)$, \(f(x,y)\) is generated by $GRF(\tau=3,\alpha=2)$ and the influence of internal or external sources/sinks \(q(x,y)\) is generated by uniform distribution $U[0,1]$.

In DiffOAS method, In the DiffOAS method, we use $50$ solution functions obtained by solving as basis functions.

\subsubsection{Noise}
We multiply the GRF function by a matrix that decays towards the edges, and then multiply it by a parameter related to the norm of the solution function to generate noise.

\subsection{Time  test}\label{time test}
In this experiment, we compare the time required to generate 10,000 data points for Darcy flow problems. The parameters used for dataset generation are consistent with Appendix~\ref{data set}. The basis functions for the DiffOAS method are generated using the GMRES algorithm with an accuracy of $1\mathrm{E}-5$, resulting in a total of 30 basis functions.

\subsection{Ablation}\label{Ablation}
In the ablation experiment, we consider the training results of 10,000 training instances generated using several different basis functions on the Darcy flow problem. The basis functions are: 
\begin{itemize}
    \item \textbf{GRF}: The function generated by the $GRF(\tau = 7, \alpha = 2.5)$ using 30 solution functions as basis functions.
    \item \textbf{Fourier}: The function generated by periodic sine wave function on the two-dimensional plane using 100 solution functions as basis functions.
    \item \textbf{Chebyshev}: The function generated by truncated Chebyshev function using 100 solution functions as basis functions.
\end{itemize}

\section{Additional Experiments}\label{Additional Experiments}

\subsection{Time comparison Experiments}\label{Time comparison Experiments}

We conducted comparative experiments involving the SKR algorithm, MINRES, and Conjugate Gradient algorithms~\citep{wang2024accelerating, paige1975solution, hestenes1952methods}. MINRES and Conjugate Gradient algorithms are often applied to specific matrix types; for instance, the Conjugate Gradient method is tailored for symmetric positive-definite matrices. To maintain consistency in presenting results, we focused on scenarios involving symmetric positive-definite matrices:

\begin{table}[ht]
\centering
\caption{
Considering the Darcy flow problem, appropriate boundary conditions are selected to ensure that the corresponding linear equation system is symmetric positive-definite. Each corresponding matrix has a size of 62500, and the table records the time taken by different algorithms to generate 1000 training data sets. The DiffOAS method's time denotes the duration for generating training data after obtaining basis functions. 
}
\label{tab:Time comparison Experiments}
\vskip 0in
\begin{center}
\begin{scriptsize}
\begin{sc}
\renewcommand{\arraystretch}{1.3}
\setlength{\tabcolsep}{20pt} 
\fontsize{9}{10}\selectfont
\begin{tabular}{cc}
\toprule
Method              & Time(s)   \\ \hline
DiffOAS(scipy)      &3.893E-1   \\ 
SKR(C)	            &1.283E3    \\ 
GMRES(scipy)	    &1.974E4    \\ 
MINRES(scipy)	    &1.149E2     \\ 
Conjugate Gradient(scipy)	&1.426E3\\
\bottomrule
\end{tabular}
\end{sc}
\end{scriptsize}
\end{center}
\vskip 0.1in
\end{table}

The experimental results show that the DiffOAS algorithm significantly outperforms other numerical algorithms in terms of computational time required for data generation. MINRES and Conjugate Gradient, optimized for specific matrix types, are faster than GMRES. Due to the large condition number of the matrices corresponding to the equations, MINRES demonstrates faster speeds than Conjugate Gradient. The SKR algorithm surpasses GMRES by optimizing the solution for multiple related linear equation systems. However, since it's not optimized for the symmetry of matrices, its speed still lags behind MINRES. Even though our method is not tailored for specific matrix types, it still generates data faster than numerically optimized algorithms (MINRES and Conjugate Gradient) designed for specific matrix types.

It is worth noting that our method can also be combined with the SKR algorithm or other numerical algorithms. Our method mainly consists of two parts: basis function generation and training data generation after obtaining the basis functions. Through the following experiments, it can be observed that the time cost of generating data with the DiffOAS algorithm after obtaining basis functions is much lower than that of all existing methods, including the SKR algorithm. Therefore, the basis function generation becomes the primary time cost of DiffOAS. The SKR algorithm significantly improves the efficiency of generating basis functions. Due to the SKR algorithm's ability to significantly enhance the efficiency of basis function generation, its integration also leads to notable performance improvement in our method.

\subsection{Error Experiments}\label{Error Experiments}

We evaluated models trained with datasets of varying accuracies generated by the GMRES algorithm to demonstrate the impact of low-precision data on model training.

The experimental results indicate that when there are significant data accuracy errors, the errors noticeably impact the performance of the model. This suggests that obtaining high-accuracy data has a significant impact on the quality of the dataset.

\begin{table}[ht]
\centering
\caption{
For the Darcy flow problem with a matrix dimension of 10,000, FNO was used as the testing model. The training outcomes of models trained on datasets generated by the GMRES algorithm, featuring truncation errors of $10^{-1}$, $10^{-2}$, and $10^{-5}$, were evaluated.
}
\label{tab:Error Experiments}
\vskip 0in
\begin{center}
\begin{scriptsize}
\begin{sc}
\renewcommand{\arraystretch}{1.3}
\setlength{\tabcolsep}{20pt} 
\fontsize{9}{10}\selectfont
\begin{tabular}{cccc}
\toprule
Train number	&tol=1E-1	&tol=1E-2	&tol=1E-5\\ \hline
100	            &1.25E-1	&4.54E-2	&4.41E-2   \\ 
500	            &1.20E-1	&1.14E-2	&6.87E-3   \\ 
1000	        &1.20E-1	&1.04E-2	&4.56E-3   \\ 
5000	        &1.20E-1	&9.60E-3	&1.69E-3   \\ 
10000	        &1.20E-1	&9.43E-3	&1.15E-3   \\
20000	        &1.20E-1	&9.85E-3	&1.01E-3   \\
\bottomrule
\end{tabular}
\end{sc}
\end{scriptsize}
\end{center}
\vskip -0.1in
\end{table}






\end{document}